\documentclass{bmvc2k}

\title{GeneGAN: Learning Object Transfiguration and Attribute
Subspace from Unpaired Data}

\addauthor{Shuchang Zhou}{shuchang.zhou@gmail.com}{1}
\addauthor{Taihong Xiao}{xiaotaihong@pku.edu.cn}{12}
\addauthor{Yi Yang}{yangyi@megvii.com}{1}
\addauthor{Dieqiao Feng}{fdq@megvii.com}{1}
\addauthor{Qinyao He}{hqy@megvii.com}{1}
\addauthor{Weiran He}{hwr@megvii.com}{1}

\addinstitution{
 Megvii Inc.\\
 Beijing, China
}
\addinstitution{
 Department of Information Science, School of Mathematical Sciences, Peking University\\
 Beijing, China
}

\runninghead{Student, Prof, Collaborator}{BMVC Author Guidelines}

\usepackage{amssymb}
\def\etal{\emph{et al}\bmvaOneDot}

\def\v{{\bf v}}

\begin{document}

\maketitle

\begin{abstract}
Object Transfiguration replaces an object in an image with another object from
a second image.
For example it can perform tasks like ``putting exactly those
eyeglasses from image A on the nose of the person in image B''.
Usage of exemplar images allows more precise specification of desired
modifications and improves the diversity of conditional image
generation.
However, previous methods that rely on feature space operations, require
paired data and/or appearance models for training or disentangling objects from
background.
In this work, we propose a model that can learn object transfiguration
from two unpaired sets of images:
one set containing images that ``have'' that kind of object, and the other set
being the opposite, with the mild constraint that the objects be located
approximately at the same place.
For example, the training data can be one set of reference face images that
have eyeglasses, and another set of images that have not, both of which
spatially aligned by face landmarks.
Despite the weak 0/1 labels, our model can learn an ``eyeglasses'' subspace that
contain multiple representatives of different types of glasses. Consequently,
we can perform fine-grained control of generated
images, like swapping the glasses in two images by swapping the
projected components in the ``eyeglasses'' subspace, to create novel images of
people wearing eyeglasses.

Overall, our deterministic generative model learns disentangled attribute
subspaces from weakly labeled data by adversarial training.
Experiments on CelebA and Multi-PIE datasets validate the effectiveness of the
proposed model on real world data, in generating images with specified
eyeglasses, smiling, hair styles, and lighting conditions etc. The code is
available online.

\end{abstract}

%-------------------------------------------------------------------------
\section{Introduction}
\label{sec:intro}
Object transfiguration is a type of conditional image generation, that first
decomposes an image into an object part and background part. The object is then
modified to satisfy a particular condition, and the background is kept unchanged.
Object transfiguration % generates an image satisfying specified condition, it
has found applications in image
editing~\cite{DBLP:journals/corr/GardnerKLUWH15,DBLP:journals/corr/BrockLRW16a,DBLP:journals/corr/PerarnauWRA16,DBLP:conf/eccv/ZhuKSE16,DBLP:journals/corr/UpchurchGBPSW16,DBLP:journals/corr/YinFSX17},
and image
synthesizing~\cite{dosovitskiy2014learning,DBLP:conf/nips/KulkarniWKT15,DBLP:conf/eccv/YanYSL16}.
Depending on the task, the object can be concrete instances like
eyeglasses, or more abstract concepts like facial expressions. 
% For example, eyeglasses have many types and models, but can still be fully
% specified if we are given an image of one person wearing the eyeglass, and 
The generated
images would then be answers to questions like ``what if her eyeglasses is
on my nose?'' or ``what if I smile like her?''

\begin{figure}
\centering
\begin{tabular}{cc}
%\bmvaHangBox{\fbox{
\includegraphics[width=0.5\textwidth]{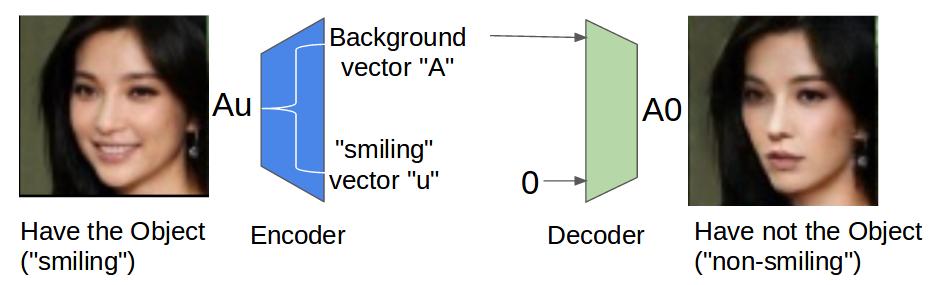} &
\includegraphics[width=0.4\textwidth]{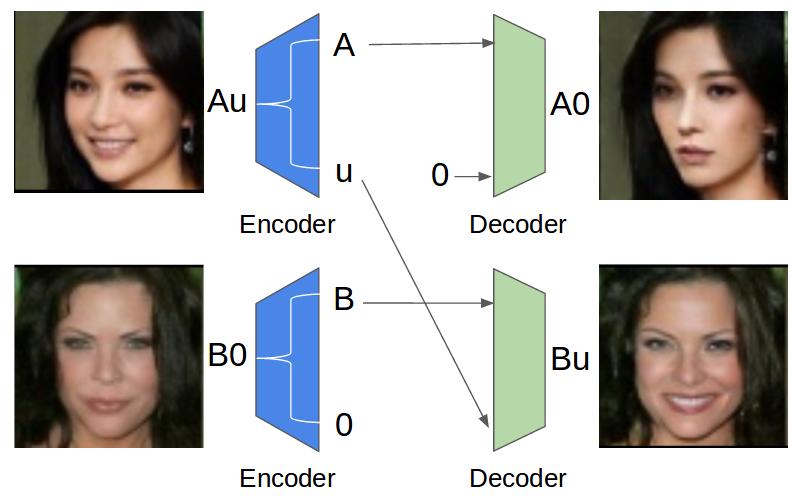} \\
(a) Object Removal & (b) Object Transplanting
%}}
\end{tabular}
\caption{(a) Encoder of GeneGAN decomposes an image to the background feature
$A$ and the object feature $u$.
The decoder can reconstruct an image without the object (a non-smiling face), from
background feature $A$ and the zero object feature (denoted as $0$).
(b) Decomposed object feature can be used to transplant the object to another
image. When the ``smiling'' feature $u$, which is from the first image $Au$, and
the background feature $B$ are fed to a decoder, the generated image $Bu$ would
ideally have the same level and style of smiling as $Au$.}
\label{fig:transform}
\end{figure}

Under the Linear Feature Space
conjecture~\cite{DBLP:conf/icml/BengioMDR13} for features extracted by Deep
Neural Networks, we may achieve complex object transfiguration tasks like
removal and transplanting of objects, by linear
operations on the feature vectors.
The images generated from the modified feature vectors can still be
natural-looking and has negligible artifacts.
% The generated images can still be natural-looking and have ghosting
% artifacts.
In fact, previous works~\cite{DBLP:journals/corr/RadfordMC15,DBLP:journals/corr/UpchurchGBPSW16}
have shown that making a face in an image smile, is as simple as addition with
an vector in feature space:
\begin{align}
\text{smiling face} = \phi^{-1}(\phi(\text{non-smiling face}) + \v_{\text{smiling}})   
\text{,}
\end{align}
where $\phi$ is a mapping from images to features, and the transform vector
$\v_{\text{smiling}}$ can be computed as the difference between clustering
centers of features of smiling faces and non-smiling faces.

However, there are many styles and levels of smiling. For example, some kinds of
smiling do not expose teeth, and some are more manifested in
eyes than in mouth. Hence representing
smiling by a single transform vector will severely limit the diversity of smiling 
in generated images. To address this diversity issue, the Visual
Analogy-Making~\cite{DBLP:conf/nips/ReedZZL15} method proposes to use a pair
of reference images to specify the transform vector, for example two images
where the same person smiles in one and not in the other.
Though this method increase the diversity, such paired data are hard to
acquire except in controlled environments.
% \begin{align}
% \v_{\text{smiling}} = \phi(\text{smiling image of person A}) -
% \phi(\text{non-smiling image of person A})
% \text{,}
% \end{align}
%However, this approach require 

%XXX

% Ideally, in 
% This approach to modification of
% the attributes of an image has attracted lots of interest due to its simpleness
% and effectiveness, and has found applications in 

% There are several challenges in learning the attribute subspace of features.
% First, it is desirable to learn a multidimensional attribute subspace that
% contains all relevant attribute vectors. Second, the attribute subspace should
% be disentangled from other factors as much as possible, to minimize the
% side-effects of modification of attributes. Third, as it is generally hard to
% collect image pairs with changes in only one attribute, it is desirable to
% design a training method that can exploit unpaired training data. Finally, the
% strength of attribute is often hard to specify, hence it would be much easier
% if we can learn from 0/1 labeled data.

Yet another approach to Object Transfiguration is the recently proposed
GAN with cyclic loss approach, which exploits Dual
Learning~\cite{DBLP:conf/nips/HeXQWYLM16,DBLP:journals/corr/KimCKLK17,DBLP:journals/corr/ZhuPIE17,DBLP:journals/corr/YiZTG17}
to map between the source images (non-smiling faces) and the target images
(smiling faces).
%for improving the stability of training. 
%An illustrative example is given in
%Figure~\ref{fig:flow}(a). 
Nevertheless, Dual Learning relies on the
invertibility of the mapping for the cyclic loss to work. When
the intrinsic dimension of the source and target domains are not the same, like
when the source domain does not have the object and the target domain has it,
the cyclic loss cannot be applied. More discussions will be left to
Section~\ref{subsec:alt-genegan}.
% However, when we remove objects, the domain will be inherently
% of lower dimension. 
% In fact, we
% observe that the images reconstructed by GAN with cyclic loss method would have
% less diversity in Section~\ref{sec:exp}.

In this work, we propose a model that can generate an object feature
vector (hereafter shortened as ``object feature'' or ``object vector'') from a
single image.
The object feature can then be transplanted to other images to generate
novel images with similar objects. Our model is made up of two parts: an Encoder
that decomposes an image to a background feature part and a object feature part, and a Decoder that can
combine a background feature and an object feature to produce an image.
Figure~\ref{fig:transform} illustrates typical usage patterns of the proposed
model: object removal and object transplanting, which are both done by
some wiring of Decoders and Encoders. All instances of Decoders and Encoders
share parameters respectively.

Moreover, object features from the Decoder are found to constitute a vector
space: the non-presence of objects is mapped to the origin, the norm is
proportional to the strength of the object (like level of smiling), and linear
combinations produce other feasible object features. Though previous
works~\cite{DBLP:journals/corr/RadfordMC15,DBLP:journals/corr/UpchurchGBPSW16}
also observe that the one-dimensional attribute vector also demonstrates these
properties, we are able to extract higher dimensional attribute subspace, due to
the increasing of diversity of object features. Experiments on face attributes
like hair styles and eyeglasses demonstrate the richness of such attribute
subspaces. We perform experiments in a proprietary Deep Learning framework, and
a Tensorflow implementation is available at \url{https://github.com/Prinsphield/GeneGAN}.

\section{Method}
\begin{figure}
\centering
\begin{tabular}{cc}
\bmvaHangBox{\fbox{\includegraphics[width=5.6cm]{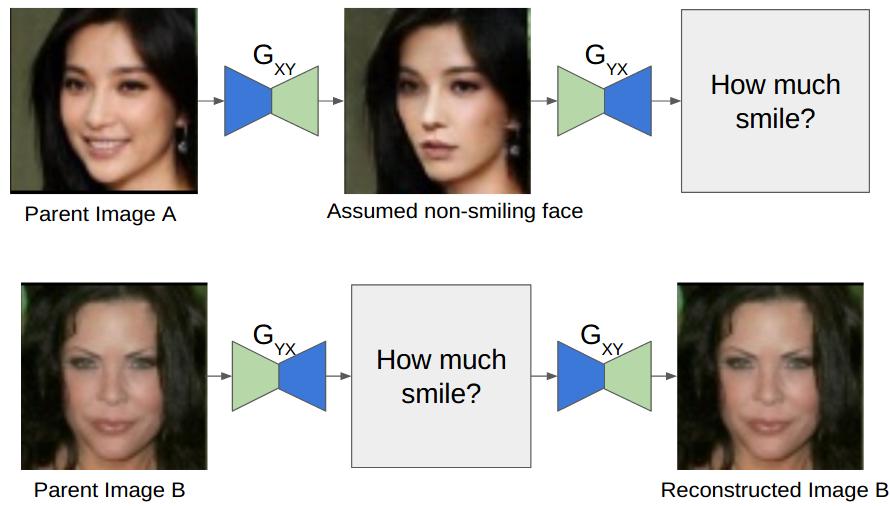}}}
&
\bmvaHangBox{\fbox{\includegraphics[width=5.6cm]{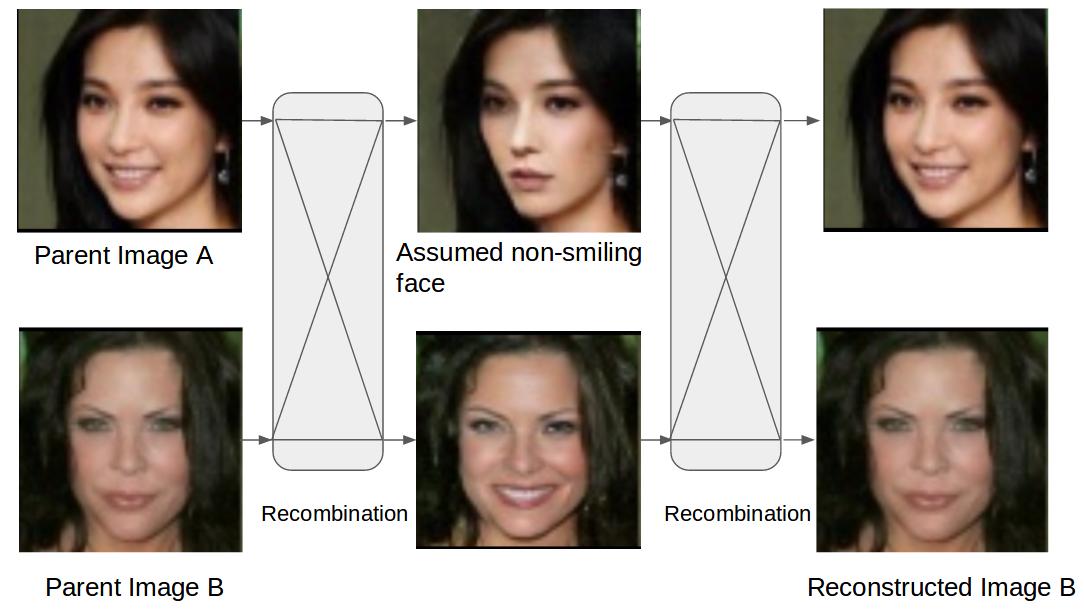}}}
\\
(a) GAN with cyclic losses &
(b) Stacked GeneGAN
\end{tabular}
\caption{
(a) The method of GAN with cyclic losses suffers from
under-determination problem when removing an object: there will be incomplete
information when going from non-smiling faces to smiling faces to determine the
level and style of smiling in generated images.
(b) The alternative Stacked GeneGAN model.
The smiling
of two images are swapped twice, through recombinations, to reconstruct the
original images.
}
\label{fig:flow}
\end{figure}

In this section, we formally outline our method and present the problem of
learning the disentangled representation for backgrounds and objects.

The training data is made of two sets: the set of image having the attributes
is $\{x_{Au}^i\}_{i=1}^N$, and the opposite set being
$\{x_{B0}^i\}_{i=1}^M$, where $u$ and $0$ stands for
presence/non-presence of an object respectively.
The two sets of images need not be paired.
The division of training data into two sets is effectively a 0/1 labeling over
all training data.

\subsection{Model}
Our model is made up of an Encoder that maps an image to two complement codes,
and a Decoder that is inverse of Encoder. Division of the codes is unknown have
to be learned from the 0/1 labeling. To achieve the disentangling of object
features from background features, we would like the following constraints to be
satisfied:
\begin{enumerate}
  \item An image without any objects should be indistinguishable from the set
  $\{x_{B0}^i\}_{i=1}^M$.
  \item An image that does have the object should be indistinguishable from the set $\{x_{Au}^i\}_{i=1}^N$.
\end{enumerate}

Here $x_{Au}$ stands for an image that will be decoded to $A$ and $u$. We will
sometimes refer to the image $x_{Au}$ simply as $Au$ when it is clear from
context.

Such ``indistinguishable'' constraint can be enforced by introduction of
adversarial discriminators~\cite{DBLP:conf/nips/GoodfellowPMXWOCB14,DBLP:journals/corr/Goodfellow17}, which
interprets indistinguishable as ``there does not exist discriminator that can
assigns different score to two sets''.

\begin{figure}
\centering
\begin{tabular}{c}
\bmvaHangBox{\fbox{\includegraphics[width=0.9\textwidth]{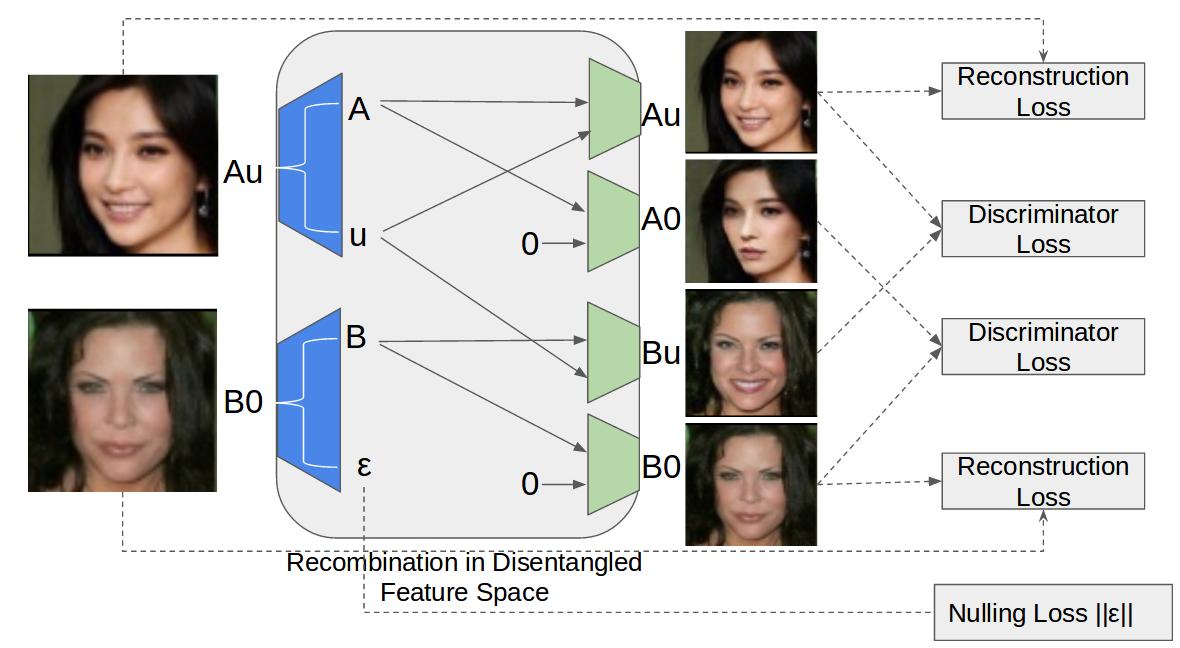}}}
\end{tabular}
\caption{The training diagram of GeneGAN. The information about the smiling in
image $Au$ is flowed to its reconstructed version and image $Bu$ through
object feature $u$.}
\label{fig:train}
\end{figure}

This inspires us to introduce the training diagram as illustrated in
Figure~\ref{fig:train} for our model, namely GeneGAN. During training, four
children $Au$, $A0$, $Bu$, $B0$ are created out of combinations of complement
codes of two parent images $Au$ and $B0$ as follows:
% Two children will be exact reconstructions of the parents and two children are
% novel recombinations that should have the same 0/1 label as one of the parents.
% The above constraints are exploited by inclusion of reconstruction losses and
% Adversarial Discriminators in training.
% 
% Our model is made up of an Encoder and the Decoder,
% which are inverse to each other. The Encoder maps an image to two complement
% codes, with one part intended to capture all latent information about
% attributes, and the other part containing all other information required to
% completely reconstruct the original image. However, since we only have 0/1
% labels about presence/non-presence of an attribute, we cannot directly specify
% how and where the attribute information is encoded. Instead, we introduce the
% following adversarial training scheme.
% 
% During training, we present the model
% with two images $Au$ and $B\epsilon$ at a time, where $Au$ has the attribute and
% $B\epsilon$ has not.
first, the Encoder will create four pieces of codes for the two images, namely
$A$, $u$, $B$ and $0$; then Decoders will create four legal
recombinations as children: $Au$, $A0$, $Bu$, $B0$.

Out of four
children, two recombinations $Au$ and $B0$ are exact reconstructions, while $A0$
and $Bu$ are novel crossbreeds. 
By using an adversarial Discriminator
to require that $Au$ being indistinguishable from $Bu$, and that we can
reconstruct $Au$ from $A$ and $u$, we can enforce all information about the
object to be encoded in $u$. Similarly, if $A0$ is not distinguishable from
$B0$, we can ensure that $A$ does not contain any information about the object. Overall, we can achieve
the disentanglement of the object information from the background information.

Moreover, the reconstruction losses will induce $u$ and $v$
to contain the complete information about the object.
% , which is much richer
% than the presence/non-presence information the training data provide. 
Inclusion of
reconstruction loss also stabilizes the training of Adversarial
Discriminator~\cite{DBLP:journals/corr/KimCKLK17,DBLP:journals/corr/ZhuPIE17,DBLP:journals/corr/YiZTG17}.
% An illustration of our method is presented in Figure~\ref{fig:flow}. The
% generated images $A\epsilon$ and $Bu$ are cross-over products of $Au$ and
% $B\epsilon$, which are novel recombinations.

% \subsection{Building Blocks}
% 
% % \begin{figure}
% % \begin{tabular}{cc}
% % \bmvaHangBox{\fbox{\includegraphics[width=6.1cm]{build-block}}}&
% % \bmvaHangBox{\fbox{\includegraphics[width=5.2cm]{cross}}}\\
% % (a)&(b)
% % \end{tabular}
% % \caption{
% % Usage patterns of GeneGAN. When a encoder }
% % \label{fig:usage}
% % \end{figure}
% 
% 
% 
% 
% It is hard to apply the simple cyclic domain translation when two
% parents belong to the same domain.
% 
% After inpainting, by definition the original image is lost. Hence there is no
% guarantee that the restored image will be the same. In fact this problem is
% inherent, and we have to resort to using a constraint.
% 
% Compared to previous method, if we are able to disentangle the factors, then we
% can precisely control how much smile to retain with the attribute part of
% feature vector.

%\subsection{Inducing a Disentangled Latent Space}

\subsection{Attribute Drift Problem and Parallelogram Constraint}
\begin{figure}
\centering
\begin{tabular}{c}
\bmvaHangBox{\fbox{\includegraphics[width=0.9\textwidth]{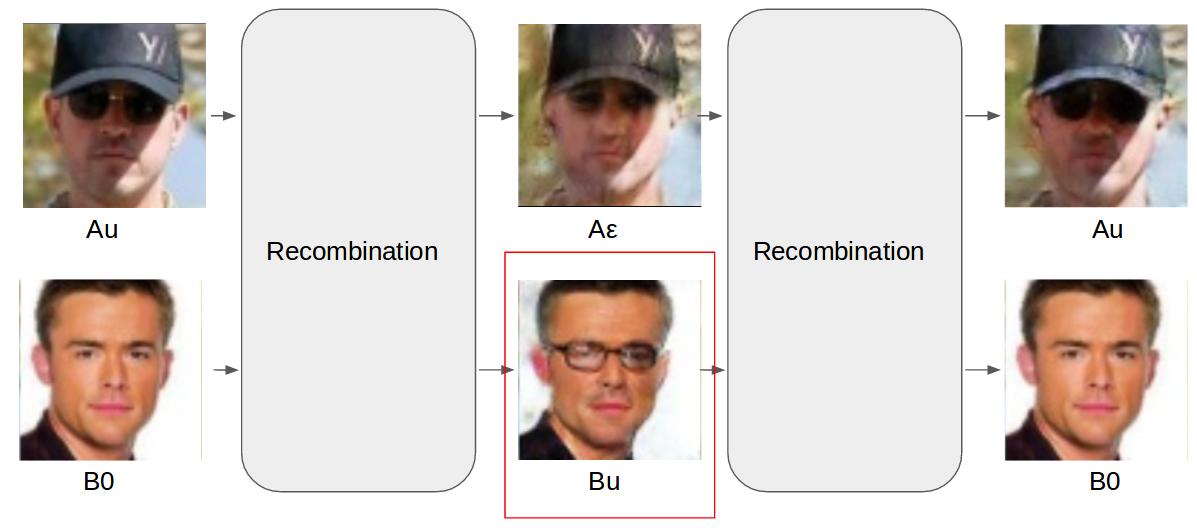}}}
\end{tabular}
\caption{The objects in $Au$ and $Bu$ may have difference appearances.}
\label{fig:attrib-drift}
\end{figure}

Through experiments, sometimes we observe a problem of ``Attribute Drift'': the
visual appearance of the objects may be different between $Au$ and $Bu$, even though
there will still be a one-to-one correspondence between the two objects.
An illustrative example is given in Figure~\ref{fig:attrib-drift}, when $Au$ is
a man wearing sunglasses with black lens and $B0$ is a man not wearing any
eyeglasses.
It is possible that $Bu$ will be the second man wearing any kind of eyeglass, as
long as there is still a one-to-one correspondence between the eyeglasses of
$Bu$ and eyeglasses of $Au$.

Instead of applying an additional classifier to enforce that styles of the
objects of $Au$ and $Bu$ be the same , we propose the following
parallelogram constraint on the image domain when objects are approximately
aligned: the sum of image pixel values of $Au$ and $B0$, should be approximately
the same as the sum of $A0$ and $Bu$. In the above example, this will encourage
a sun glass, when transplanted, to stay as a sun glass, and not mutating into an
eyeglass.

Note it will not make sense to include the parallelogram loss for GAN with
cyclic losses, as the transformation of two original images are completely independent
of each other. In their case, adding a parallelogram loss will only increase the
overfitting level of the model.

\subsection{Loss Function of Training}
Given two images $x_{Au}$ and $x_{B0}$, the data flow of training of
GeneGAN can be summarized in following equations:
\begin{align}
\begin{array}{cc}
(A, u) = \operatorname{Encoder}(x_{Au}) & (B, \epsilon) =
\operatorname{Encoder}(x_{B0})\\
x_{A0} = \operatorname{Decoder}(A,0) &x_{Bu} =
\operatorname{Decoder}(B, u) \\
x_{Au}' = \operatorname{Decoder}(A, u) & x_{B0}' =
\operatorname{Decoder}(B,0) \\
\end{array}
\text{.}
\end{align}
We force the $\epsilon$ encoded from $B0$ to be zero, so as to ensure the
constraint that $A0$ should not contain any information about $B0$, and
that any information contained in the object part of $B0$ can be safely
discarded.

The generator receives four types of losses: (1) the standard GAN losses, which
measures how realistic the generated images are; (2) the reconstruction losses,
which measures how well the original input is reconstructed after a sequence of encoding and decoding. (3) the 
nulling losses, which reflects how well the object features are disentangled
from background features and (4) (optionally) the parallelogram losses, which
enforces a constraint between the children and the parents in image pixel
values.
We omit the weights of the losses and left the details to online implementation.
$P_0$ and $P_{\neq 0}$ stand for the distribution of images without and with the
objects, respectively.

\begin{align}
\begin{array}{cc}
L_{reconstruct}^{Au} = \|x_{Au} - x_{Au}'\|_1 & 
L_{reconstruct}^{B0} = \|x_{B0}-x_{B0}'\|_1
\\
L_{GAN}^{0} = -\mathbb{E}_{z \sim P_0}[\log D(x_{A0}, z)] &
L_{GAN}^{\neq0} = -\mathbb{E}_{z \sim P_{\neq0}}[\log D(x_{Bu}, z)]
\\
L_{0} = \| \epsilon\|_1 & 
L_{parallelogram} = \|x_{Au} + x_{B0} - x_{A0} - x_{Bu}\|_1 \\
\multicolumn{2}{c}{
L_G = L_{reconstruct}^{Au} + L_{reconstruct}^{B0} + 
		L_{GAN}^{0} + L_{GAN}^{\neq0} + 
		L_{0} + L_{parallelogram}} 
\end{array}
\end{align}

The nulling loss will push the background information to the $B$ part. In fact,
the object output of encoder will not contain background information, as
$\epsilon$ is forced to zero.

The Reconstruction losses will serve multiple purposes. First, the
reconstruction losses associated with $Au$ and $B0$ will ensure that Decoder and
Encoder are inverse to each other. In addition, $A$ is forced to contain background
information, to allow Decoder to reconstruct $Au$ from $A$ and $u$.

The discriminator receives the standard GAN discriminator loss 
\begin{align}
\begin{split}
L_{D}^{0} &= - \mathbb{E}_{z \sim P_0}[\log D(x_{Au}, z)] - \mathbb{E}_{z \sim P_{0}}[\log(1- D(x_{Bu}, z))]\\
L_{D}^{\neq0} &= - \mathbb{E}_{z \sim P_{\neq0}}[\log D(x_{A0}, z)] - \mathbb{E}_{z \sim P_{\neq0}}[\log(1- D(x_{B0}, z))]\\
L_{D} &= L_{D}^{0} + L_{D}^{\neq0} 
\end{split}
\end{align}

% \begin{align}
% L_{GAN} = L_{GAN}^{\epsilon} + L_{GAN}^{\neq\epsilon}\\
% L_{reconstruct} = L_{reconstruct}^{Ac} +
% L_{reconstruct}^{B\epsilon} \\
% L_{total} = L_{GAN} + \gamma_1 L_{reconstruct} + \gamma_2 L_{\epsilon} +
% \gamma_3 L_{parallelogram}
% \end{align}

The Discriminator losses will ensure the recombinations be of desired property.
% impose constraints on the learned codes. Take the
% smiling as an example. The Discriminator losses will ensure that the background
% output of encoder will not contain smiling information, as $B\epsilon$ is not smiling. $u$ 
% will have to contain the smiling information, as $Bu$ is a smiling face.

As we enforce constraints by losses, the constraints hold only approximately
and there will be potential leakage of information between the object and
feature parts. We leave it as future work to explore even stronger enforcement
of constraints.

% \subsection{Object Subspace}
% Cyclic loss is good. However, being able to fully reconstruct the parent images
% requires a complete latent space. If we only use an attribute vector, factors
% not captured by labeling, for example the fine grained model of the glasses,
% will be lost, and there is no guarantee that the same glasses will be put back
% to the original person.
% 
% On the other hand, the glass information will leak to the other part, destroying
% the disentangling property.
% 
% It is hard to fully characterize the latent space.

% \subsection{Avoiding the Mode Collapse with Attributes Recombination}
% GAN are known to often incur mode collapse as there are no terms to directly
% penalize non-diversity of the output.
% As GeneGAN allows combination attributes of two parent images, it will create
% interesting combinations.
% 
% Moreover, we can create hybrids by picking points in attribute subspace.

\subsection{Alternative Stacked GeneGAN}
\label{subsec:alt-genegan}
Method of GAN with cyclic loss suffers from under-determination problem when
performing object removal and reconstruction.
For example, in Figure~\ref{fig:flow}(a), though the non-smiling version of a
smiling face is well defined, it is hard to determine how much smiling should be
present when mapping from the non-smiling faces to smiling faces. 

However, it is also possible to add cross-links to allow communication of
information.
In fact, we have also experimented with a training diagram closer to the method
of GAN with cyclic losses, as illustrated in Figure~\ref{fig:flow}. The difference
between this and Figure~\ref{fig:train} is that only crossbreeds
are created as children of the parent images.
The children will be recombined again to produce grand-children, which ideally have the same features
as their grand-parents. Reconstruction losses can then be used to enforce this
invariance, exactly as is done in the method of GAN with cyclic losses.

However, we find this ``double-swap'' training diagram to be inferior in
experiments. We observe that the reconstruction losses will be significantly
higher as the grand-children went through more nonlinear transformations. We
conjecture that the higher reconstruction losses will compete more with the
adversarial losses, and create instability of training and degrades the quality of generated images.

\section{Experiments}
\label{sec:exp}
In this section we perform experiments on real-world datasets to validate the effectiveness of our method.
For training, we used learning rate of 5e-5 and momentum of 0, under
RMSProp~\cite{DBLP:conf/icml/SutskeverMDH13} learning rule. The
Neural Network models used in this section are all equipped with Batch
Normalization~\cite{ioffe2015batch} to speedup convergence. The encoders have
three layers of convolutions with Leaky ReLU
nonlinearity~\cite{maas2013rectifier}. The decoders have
three convolution layers with fractional
stride~\cite{DBLP:journals/corr/SpringenbergDBR14}.
More details of the models will be available online.

Experiments are done on Linux
machines with Intel Xeon CPUs and NVidia TitanX Graphic Processing Units.

\subsection{Dataset}
The CelebA~\cite{DBLP:conf/cvpr/LiuLQWT16} dataset is a large-scale face
attributes dataset including 202599 face images of 10177 identities, each with 40 attributes
annotations and 5 landmark locations. The landmark locations can be used to
spatially align the faces.

The Multi-PIE database~\cite{moore2010multi} contains over 754,200 images from
337 subjects, captured under 15 viewpoints and 19 illumination conditions.

%\subsection{Annulling and Replacing of Objects}

\subsection{Swapping of Attributes}
\begin{figure}
\centering
\begin{tabular}{c}
%\bmvaHangBox{\fbox{
\includegraphics[width=0.9\textwidth]{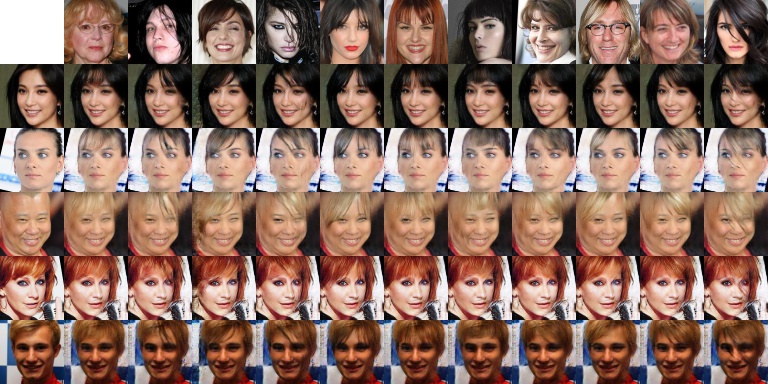} \\
(a) hair style \\
\includegraphics[width=0.9\textwidth]{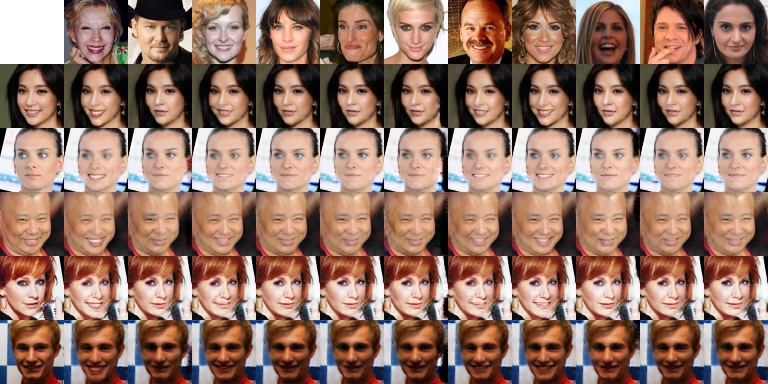} \\
(b) smiling\\
\includegraphics[width=0.9\textwidth]{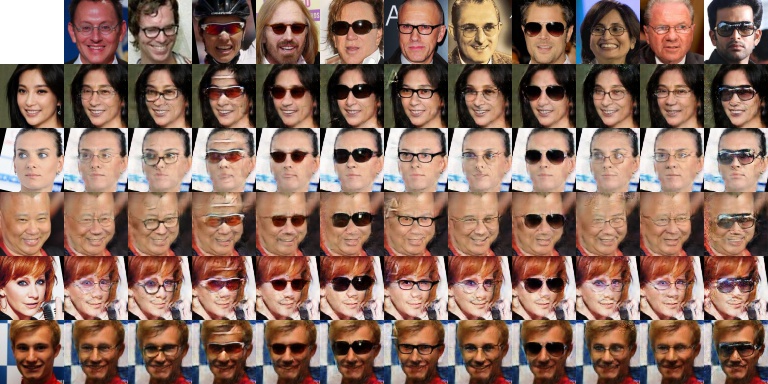} \\
(c) glasses
%}}
\end{tabular}
\caption{Replacing the object for images in each row with those of the
column heads.}
\label{fig:smiling}
\end{figure}

% \begin{figure}
% \begin{tabular}{c}
% %\bmvaHangBox{\fbox{
% 
% %}}
% \end{tabular}
% \caption{Replacing the hairs in images with those of the column heads.}
% \label{fig:hair}
% \end{figure}
% 
% 
% \begin{figure}
% \begin{tabular}{c}
% %\bmvaHangBox{\fbox{
% 
% %}}
% \end{tabular}
% \caption{Replacing the eyeglasses in images with those of the column heads.
% }
% \label{fig:eyeglasses}
% \end{figure}

%Here we compare with GAN with cyclic loss.
As GeneGAN only exploits the weak 0/1 labels of the images, and that there
will be no training data about the recombined versions, we will not distinguish
between the training data and test data in experiments carried out in this
subsection.

An object can be removed by passing in $\epsilon$ to the Decoder. Similarly,
an object can be replace by replacing the object input to the decoder.
As our model can disentangle the object part from background part, the
crossing usage pattern as illustrated in Figure~\ref{fig:transform}(b) can also be
used to swap the attributes.

In each row of the following diagram, the object parts of the images are
overridden by the objects of the first image in the row. Of particular interest
is Figure~\ref{fig:smiling}(a). It can be observed that the hair styles follow
closely the source images on the top row. In fact, the directions of hairs is in good
agreement with the source images.

\begin{figure}
\centering
\begin{tabular}{cc}
\includegraphics[width=0.4\textwidth]{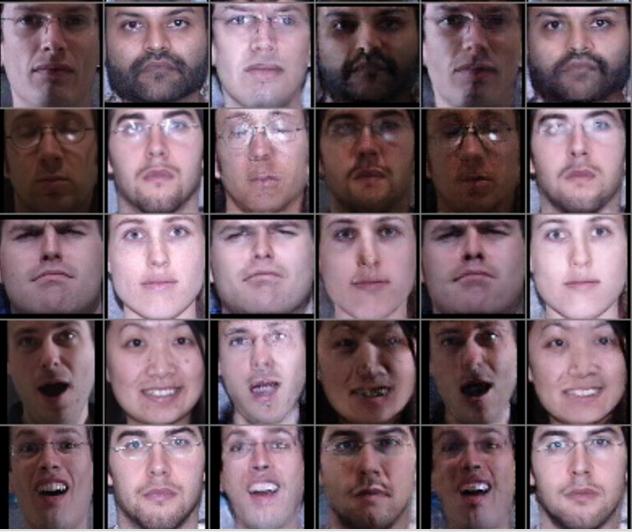} &
\includegraphics[width=0.4\textwidth]{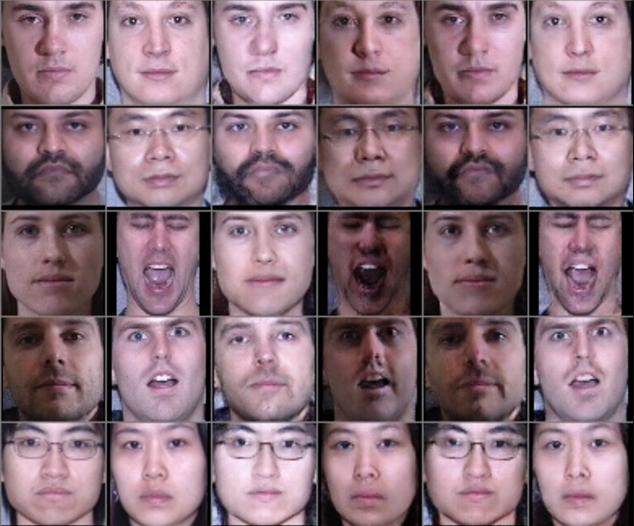} \\
\end{tabular}
\caption{Swapping the lighting conditions of two faces on Multi-PIE dataset.
From left to right, the six images in a row are: original $Au$ and $B\epsilon$,
recombined $A\epsilon$ and $Bu$, and reconstructed $Au$ and $B\epsilon$ respectively.}
\label{fig:lighting}
\end{figure}

\subsection{Generalization to Unseen Images and Comparisons with GAN with
cyclic losses}
\begin{figure}
\centering
\begin{tabular}{cc}
\includegraphics[width=0.43\textwidth]{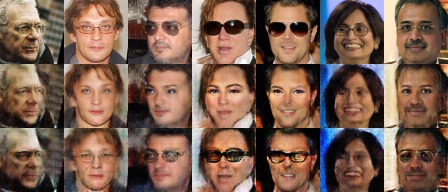}  &
\includegraphics[width=0.43\textwidth]{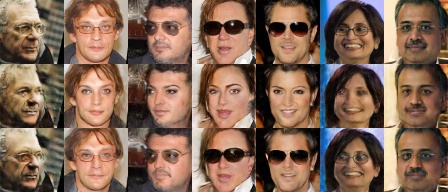} \\
(a) GAN with cyclic losses & (b) GeneGAN\\
\end{tabular}
\caption{Comparison between GAN with cyclic losses (the best model before
divergence), and GeneGAN.
The top row, middle row, and the bottom row are images with original,
removed and reconstructed objects respectively.}
\label{fig:disco}
\end{figure}
% \subfloat[GAN with cyclic loss]{\includegraphics[width=0.45\textwidth]{disco-gan}}
% \subfloat[GeneGAN]{\includegraphics[width=0.45\textwidth]{gene-gan-test}}

Figure~\ref{fig:disco} compares GeneGAN with DiscoGAN. Though DiscoGAN may also
reconstruct images that are of good quality, the objects in the reconstructed
images are not quite related to the original images. In contrast, GeneGAN
produces consistent reconstruction.

\begin{figure}
\centering
\begin{tabular}{c}
\includegraphics[width=0.7\textwidth]{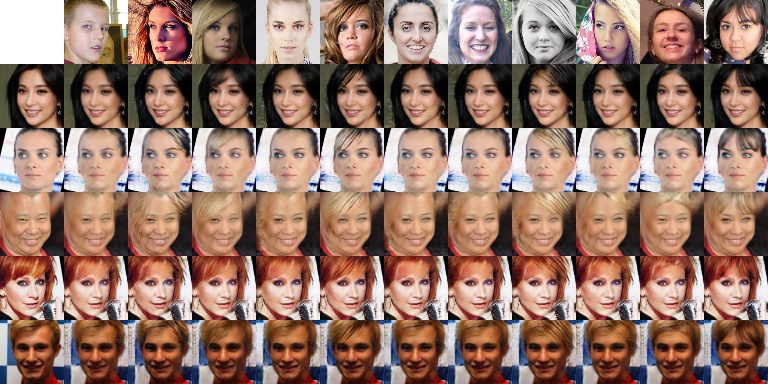}
\end{tabular}
\caption{Performance of GeneGAN on unseen data from Wider
Face~\cite{DBLP:conf/cvpr/YangLLT16}.}
\label{fig:unseen}
\end{figure}

Figure~\ref{fig:unseen} gives results of testing a GeneGAN model trained on
CelebA dataset on the Wider Face dataset, which contains face images in even
less constrained environments. It can be seen that the GeneGAN model generalizes
well to unseen data.

% \subsection{Evaluation whether children are of the same distribution as parents}
% Evaluation of generative models is
% difficult~\cite{DBLP:journals/corr/TheisOB15}.

% \subsection{Evaluation of Diversity}
% We adopt MODE~\cite{DBLP:journals/corr/CheLJBL16} score to evaluate output
% quality and diversity.

\subsection{Interpolation in Attribute Subspace}
\begin{figure}
\centering
\begin{tabular}{cc}
\includegraphics[width=0.4\textwidth]{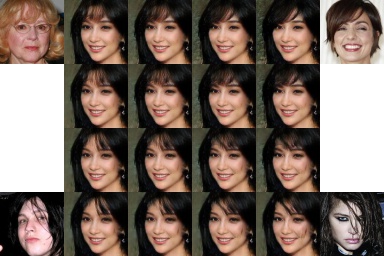}  &
\includegraphics[width=0.4\textwidth]{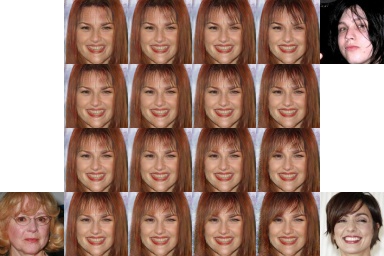} \\
\end{tabular}
\caption{The attribute space as interpolated by several object feature vectors.}
\label{fig:interpolation}
\end{figure}

Figure~\ref{fig:interpolation} gives interpolation of object features in
attribute subspace. Note the backgrounds (human identities) are approximately
the same, while objects (hair styles) are interpolated.

\section{Related Work}
In this section we discuss related work not yet covered.

Using autoencoders for transforming images
may be traced back to Hinton \etal~\cite{DBLP:conf/icann/HintonKW11}.
When one of the disentangled space is known, like having class labels,
techniques like statistical independence can be used to disentangle the two
spaces~\cite{DBLP:conf/nips/KingmaMRW14,DBLP:journals/corr/CheungLBO14,DBLP:journals/corr/LouizosSLWZ15,DBLP:conf/nips/ReedZZL15}.
In this work, we do not assume that any of the two spaces have additional
supervision signals.

\section{Conclusion}
This paper presents GeneGAN, a deterministic conditional generative model that
can perform object transfiguration task, which is modification of an object in an image with
background unchanged.
The proposed model learns to disentangle the object features from other factors
in feature space from weakly supervised 0/1 labeling of training data.
Consequently, our model can extract an object feature vector from a single image
and transplanted it to another image, hence allows fine-grained control of
generated images, like ``putting eyeglasses of A onto noses of B''. The objects
can be abstract and difficult to characterize, like hair styles and
lighting conditions.
The training method for our model is symmetric and allows exploiting cyclic
reconstruction loss, which improves stability of training. 

% Moreover, the
% crossbreeding between parents images shortens the path from the parent images to
% their reconstruction, which is also found to be beneficial to the stability of
% training.
%We also show that parallelogram loss helps.
The setup of our model also gives rise to an attribute subspace, which contains multiple vectors that are representatives of
different objects. The vectors can be scaled, inverted and interpolated to
manipulate the objects in generated images.
% We also analyze the
% attribute subspace with matrix decomposition methods, and is able to find representative vectors that caputre typical patterns of
% attributes. Like ``eigen-face'', we are able to define ``eigen-smile''.

As future work, it would be interesting to investigate whether more complex
crossbreeding patterns between more parents would allow further improvement of
stability of training, quality and diversity of generated images.

\bibliographystyle{bmvc2k_natbib}
\bibliography{thesis}
\end{document}